\documentclass[a4paper, 10pt, conference]{templates/ieeeconf}      

\IEEEoverridecommandlockouts                              

\usepackage[whole]{bxcjkjatype} 
\usepackage{xcolor}
\usepackage{graphicx}
\usepackage{tabularx} 
\usepackage{amssymb} 
\usepackage{booktabs}
\usepackage{hyperref}

\title{\LARGE \bf
On-Demand Human Assistance for Task Continuation under \\Physical Action Failures in LLM-based Planning
}

\author{Shoichi Hasegawa$^{1,*}$,
        Akira Taniguchi$^{2, 1}$,
        Lotfi El Hafi$^{3, 4}$,\\
        Gustavo Alfonso Garcia Ricardez$^{5, 6}$,
        Tadahiro Taniguchi$^{7, 4}$
        \thanks{
            This work was supported by JSPS KAKENHI Grants-in-Aid for Scientific Research (Grant Numbers JP23K16975) and JST Moonshot Research \& Development Program (Grant Number JPMJMS2011).
        }
        \thanks{
            $^{1}$Shoichi Hasegawa and Akira Taniguchi are with Ritsumeikan University;2-150 Iwakura, Ibaraki, Osaka 567-8570, Japan.
            $^{2}$Akira Taniguchi is with Kyushu Institute of Technology;680-4 Kawadu, Iizuka, Fukuoka 820-0067, Japan.
            $^{3}$Lotfi El Hafi is with Coarobo GK;2-2-2 Hikaridai, Seika, Soura, Kyoto 619-0237, Japan.
            $^{4}$Lotfi El Hafi and Tadahiro Taniguchi are with Ritsumeikan University Research Organization of Science and Technology;1-1-1 Nojihigashi, Kusatsu, Shiga 525-8527, Japan.
            $^{5}$Gustavo Alfonso Garcia Ricardez is with The University of Osaka;1-3 Machikaneyama, Toyonaka, Osaka 560-8531, Japan.
            $^{6}$Gustavo Alfonso Garcia Ricardez is with Ritsumeikan Global Innovation Research Organization;1-1-1 Nojihigashi, Kusatsu, Shiga 525-8527, Japan.
            $^{7}$Tadahiro Taniguchi is with Kyoto University;
            Yoshida Honmachi, Sakyo, Kyoto, Kyoto 606-8317, Japan.
            {\tt\small\{hasegawa.shoichi, lotfi.elhafi\} @em.ci.ritsumei.ac.jp}, 
            {\tt\small taniguchi.akira255@mail.kyutech.jp},
            {\tt\small garcia.g.es@osaka-u.ac.jp},\protect\\
            {\tt\small taniguchi@i.kyoto-u.ac.jp}
        }%
        \thanks{
            $^{*}$Corresponding author.
        }%
}

\begin{document}


\maketitle


\thispagestyle{empty}
\pagestyle{empty}


\begin{abstract}
While robot action planning based on large language models (LLMs) has advanced remarkably, continuing a task after a physical action failure during execution remains a key challenge.
For example, after failing to grasp an object, the robot may not notice that it has fallen off the table and keep trying to detect and pick it up, stalling the task.
Existing approaches either rely on closed-loop autonomous re-planning, which fails when errors exceed the robot's capabilities, or on human intervention without updating the LLM's plan, hindering subsequent planning.
We present a system for LLM-based robot action planning that integrates remote human intervention with feedback-based replanning.
When the LLM detects an action failure, the robot requests remote assistance; an operator resolves the failure through teleoperation and reports the outcome in natural language, which is fed back to the LLM to update its plan so that the task can continue.
We evaluated the proposed system on a real-world trash-collection task, comparing it with a fully autonomous method and a teleoperation-only baseline.
The proposed system improves task progress over the autonomous method in this case study.
For easy-to-retrieve trash, the gap from the teleoperation reference was small, while a larger gap remained for harder-to-retrieve trash.
These results suggest that, in this case study, combining remote human intervention with feedback-based replanning was associated with continued task execution even in the face of otherwise unrecoverable failures.
The project website is \href{https://emergentsystemlabstudent.github.io/REPAIR/}{\textcolor{blue}{https://emergentsystemlabstudent.github.io/REPAIR/}}.
\end{abstract}


\section{Introduction}
\label{sec:introduction}
With the technological advancement of large language models (LLMs), research on using LLMs to generate action sequences for robots has progressed remarkably~\cite{zhai2025survey}. 
Such LLM-based action planning has a broad range of applications, and studies have been conducted in diverse environments such as domestic settings~\cite{ahn2022can} and factories~\cite{tsushima2025task}.
These methods have enabled LLMs to generate robot action plans from a wide variety of natural-language instructions. 
However, handling physical action failures that occur during execution, such as failures in object grasping or object detection, remains a challenge.

\begin{figure}[t]
    \centering
    \includegraphics[width=1.0\linewidth]{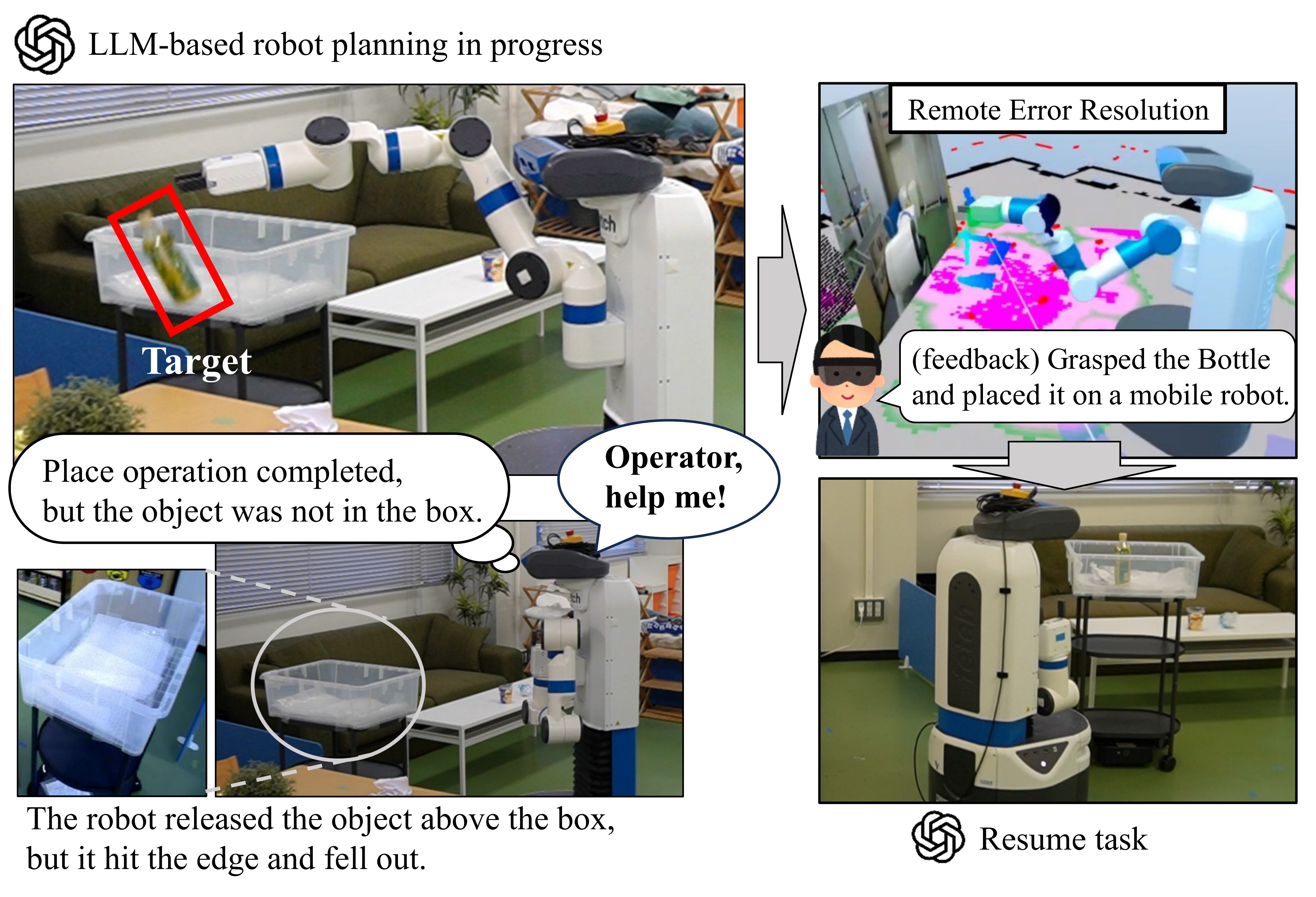}
    \caption{
        An overview of our study. When a robot drops an object and subsequently fails to detect it, it falls into a loop state. In such cases, the robot requests assistance from an operator on demand, and the issue is resolved through remote error resolution.
    }
    \label{fig:abstract}
\end{figure}

To address this challenge, research has so far proceeded in two broad directions. 
The first is a direction in which the system autonomously re-plans in a closed-loop manner~\cite{huang2022inner,liu2023reflect}, while the second is one in which a human intervenes during execution to recover from failures~\cite{ly2026inteliplan,liu2023llm}. 
However, the former does not involve any physical human engagement, and therefore cannot escape the failure loop when a physical failure exceeds the robot's capabilities. 
The latter faces limitations as well. 
Methods based solely on verbal instructions still require the robot itself to perform the physical action, and thus fall into the same limitation. 
In contrast, methods in which a human physically acts on the robot's behalf fail to reflect the resulting changes in the LLM's plan, which hinders subsequent planning even when the intervention has altered the situation.
Therefore, both a mechanism for physical human intervention and one for reflecting its outcome in the LLM's plan to continue autonomous execution are required.

Such human intervention in a robot's autonomous execution has been studied extensively as human-in-the-loop (HITL) systems. 
Among these, remote error resolution (RER) has a remote operator intervene to resolve failures that occur during execution~\cite{kumar2025mixed,kaipa2015resolving}. 
However, existing RER focuses on resolving the failure itself and does not reflect the outcome of the intervention back into the LLM's plan, leaving a discrepancy between the plan and the environment as autonomous execution continues. 
Thus, a framework that integrates physical human intervention with reflecting that outcome in the LLM's plan has not yet been realized.

Therefore, this study aims to enable the continuation of task execution in LLM-based robot action planning by, when the robot fails at a physical action, resolving the failure through remote intervention and feeding the outcome back into the LLM's plan. 
To this end, we build a system that integrates LLM-based action planning, RER, and re-planning that reflects the outcome of the intervention. 
Fig.~\ref{fig:abstract} shows an overview of this study. 
During task execution, the robot may fall into a loop of repeating the same action due to a physical failure. 
In the proposed system, when an action such as object detection or picking fails, the LLM requests assistance from the operator.
The operator intervenes remotely and feeds the outcome back, allowing the robot to resume the task.

The contributions of this study are two-fold.
\begin{enumerate}
    \item We present a case study of a human-in-the-loop system that detects physical action failures during the execution of LLM-generated plans, requests human assistance on demand, and resumes planning by feeding the resolved state back to the LLM.
    \item Through a real-world trash-collection experiment, we show that on-demand human assistance improves task progress over full autonomy in this setting. Its effectiveness, however, may depend on the failure characteristics: descriptively, it approached the teleoperation reference for Level~1 objects, but a larger gap remained for Level~2 objects.
\end{enumerate}


The remainder of this paper is organized as follows.
Section~\ref{sec:related_works} reviews the related work.
Section~\ref{sec:proposed_method} describes the proposed system.
Section~\ref{sec:experiment} presents the experiment.
Section~\ref{sec:discussion} presents the discussion about the experiment.
Section~\ref{sec:limitation} discusses the limitations of this study.
Finally, Section~\ref{sec:conclusion} concludes the paper and outlines future work.

\section{Related Work}
\label{sec:related_works}

\subsection{LLM-Based Action Planning and Its Challenges}

Many studies have provided LLMs with a robot's skill set and task-solving strategies via few-shot prompting, letting the LLM generate robot action sequences sequentially~\cite{zhai2025survey}. 
A representative example is SayCan~\cite{ahn2022can}. 
Given a language instruction, it selects actions based on both their linguistic likelihood and their estimated success probability from the current state.
Actions are then executed in descending order of this value. 
Many subsequent methods build on SayCan. 
These include approaches that enrich the environmental information referenced during action planning~\cite{hasegawa2025spatial}, and approaches that focus on program generation in addition to action planning~\cite{liang2023code}.

Among these, recovery from physical action failures that arise during LLM-based action planning and execution is an important challenge. 
Research has so far proceeded in two broad directions. 
The first is a direction in which the system autonomously re-plans in a closed-loop manner. 
Inner Monologue feeds several types of information back to the LLM as feedback, such as task success or failure, the list of detected objects, and scene descriptions representing progress. 
This feedback prompts the LLM to re-plan its actions~\cite{huang2022inner}. 
Other work explains the cause of failure and re-plans based on a summary of observations obtained after task execution~\cite{liu2023reflect}. 
Another method has the system itself search for the information it needs according to the type of failure, revise the prompt, and recover autonomously~\cite{shirasaka2024self}.
The second is a direction in which a human intervenes during action execution to recover from failures. 
InteLiPlan has the robot explain the reason for the failure to a human. 
The human then provides additional instructions that prompt the LLM to re-plan~\cite{ly2026inteliplan}. 
In other work, rather than having a human give verbal instructions, a human teleoperates the robot when an action fails. 
The recorded trajectory is then reproduced using Dynamic Movement Primitives, complementing the shortcomings of the LLM's autonomous execution~\cite{liu2023llm}.

Closed-loop systems re-plan without physical human involvement, and thus cannot recover from failures that exceed the physical capabilities of the robot.
Approaches that involve a human do not reflect the outcome of the physical intervention back into the LLM's plan, causing a mismatch between the plan and the environment in subsequent execution.
In contrast, this study integrates physical human intervention with a mechanism that reflects the intervention's outcome in the LLM's plan, enabling autonomous execution to continue.

    \begin{figure*}[t]
        \centering
        \includegraphics[width=1.0\linewidth]{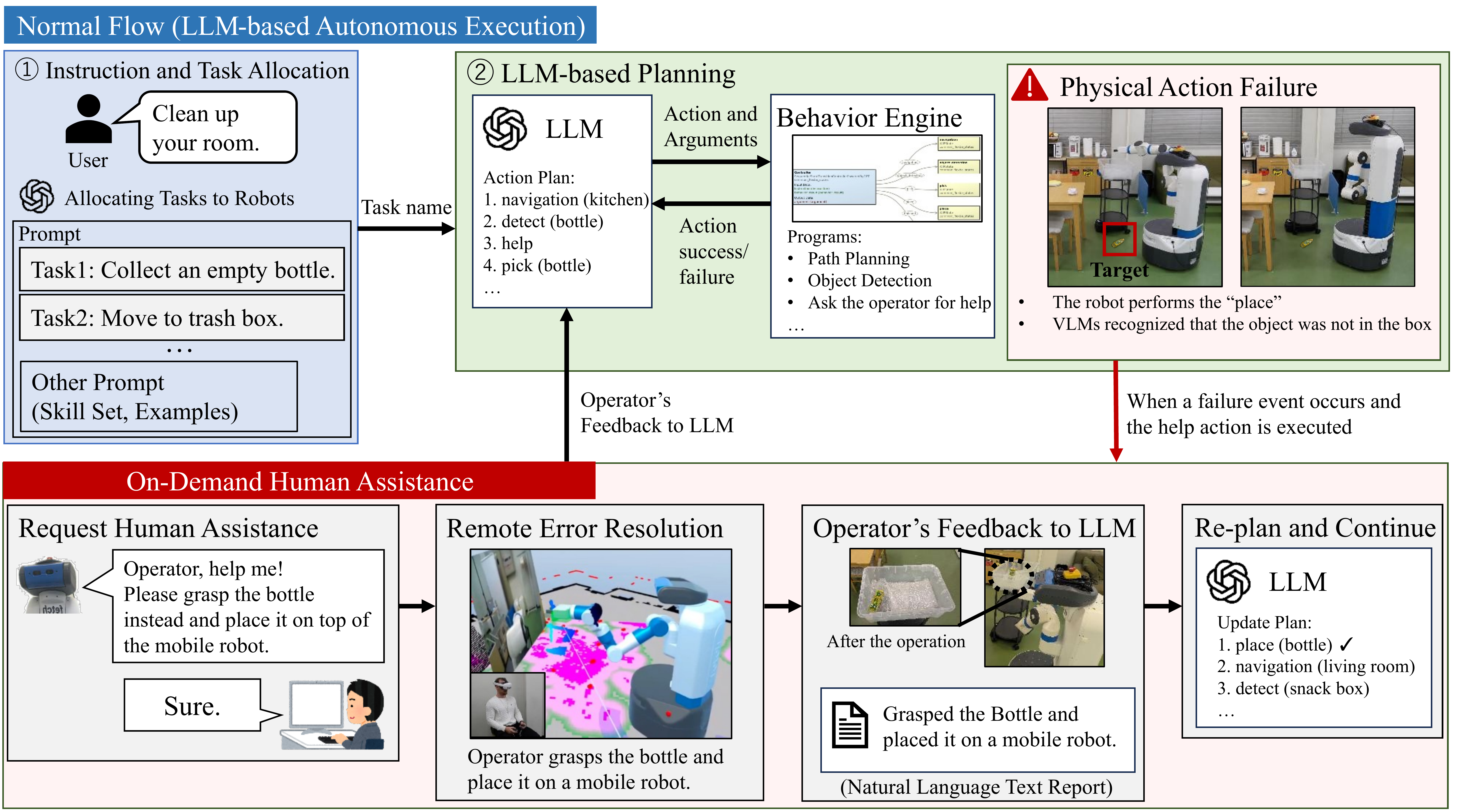}
        \caption{
            An overview of the proposed system.
            The LLM takes a user's instruction, decomposes it based on robot capabilities and rules, and assigns tasks to each robot. After that,  LLMs generate executable actions, which the robots perform.
            When failures occur, a robot requests operator assistance. The operator resolves the issue remotely and provides feedback, allowing the LLM to update its understanding and resume execution.
        }
        \label{fig:proposed_method}
    \end{figure*}

\subsection{Human Intervention in Robot Task Execution and Its Challenges}

Research on HITL has aimed to improve the quality and reliability of a robot's actions. 
It does so by incorporating human judgment and operation into the robot's autonomous execution. 
Among these, Shared Autonomy / Shared Control is one framework. 
Here, humans and robots leverage their complementary strengths to achieve a common goal: the robot handles tasks that can be automated, while the human operates the robot to handle the rest~\cite{hagenow2025shared}. 
For example, Heger~\textit{et~al.} propose Sliding Autonomy (Adjustable Autonomy)~\cite{Heger2006sliding}. 
In this method, a task is decomposed into fine-grained subtasks with a defined execution order. 
For each subtask, control is switched between the autonomous robot and the operator depending on the situation. 
In a structure-assembly experiment, they report improvements in both efficiency and reliability.
Other work goes a step further. 
It not only lets a human intervene on the spot but also uses the operation data from the intervention for policy learning, improving performance on contact-rich tasks~\cite{liu2025robot}. 
Within this line of work, the framework called RER is particularly relevant. 
In RER, a remote operator intervenes to resolve the problem when a failure occurs during a robot's autonomous execution. 
Kumar~\textit{et~al.} apply this method in a task that emulates warehouse pick-and-place~\cite{kumar2025mixed}. 
There, a single robot arm makes perception or manipulation errors. 
As a result, they achieve shorter task-completion times and improved usability compared with camera- and VR-based methods. 
Wozniak~\textit{et~al.} and Kaipa~\textit{et~al.} focus on perception errors in the robot's object detection. 
They build mechanisms in which the operator corrects the errors using VR~\cite{wozniak2023happily,kaipa2015resolving}.

Existing RER focuses primarily on resolving failures through intervention. 
It does not provide a mechanism for reflecting the changes caused by the intervention in subsequent planning. 
As a result, even if a single failure is resolved, a discrepancy arises later. 
The plan and the real environment diverge as autonomous execution continues. 
To the best of our knowledge, prior RER studies have primarily focused on resolving the immediate failure rather than explicitly incorporating the intervention outcome into subsequent LLM-based planning.
Therefore, this study proposes a method that integrates RER into LLM-based action planning. 
The outcome of the operator's intervention is reflected in the plan as feedback to the LLM. 
In this way, autonomous execution continues even after a failure is resolved. 
A distinctive feature concerns how the intervention is controlled. Prior work used a state machine, whereas our method has the LLM handle everything from triggering the intervention to re-planning.

\section{Proposed System}
\label{sec:proposed_method}
In this study, we propose a system for LLM-based robot action planning that continues a task by resolving physical action failures through remote human intervention (Fig.~\ref{fig:proposed_method}). 
When the robot fails at a physical action, a human intervenes remotely to resolve the failure. 
The outcome of the intervention is then reflected in the LLM's plan, so that the task can continue. 
The proposed system consists of two components. 
The first is LLM-based task decomposition, assignment, and action planning. 
The second is a mechanism for handling physical action failures: when the robot's action fails, the operator intervenes remotely (RER), and the outcome is fed back to enable re-planning.

\subsection{LLM-Based Action Planning}
\label{subsec:proposed_action_planning}

Our method targets a system in which one or more robots execute action plans generated by an LLM. 
First, the LLM receives an instruction from the user and, based on a prompt for task decomposition and assignment, decomposes the instruction into a form that the robots can execute. 
It then assigns each subtask to a robot according to its skills. 
The prompt for task decomposition and assignment consists of the following elements:
\begin{enumerate}
    \item description of the task
    \item physical information of each robot
    \item names of the rooms in the environment
    \item placement locations of objects
    \item rules for task decomposition and assignment
    \item examples of task decomposition and assignment
    \item user instruction
\end{enumerate}
The prompt used for task decomposition and assignment was constructed based on our prior work~\cite{murata2025multi}.

For each assigned subtask, the LLM generates an action plan using a dedicated prompt prepared for the robot. 
This prompt includes the room names, object locations, the action skills available to the robot, the planning rules, and examples.
In this study, the robots can use the following action skills:
\begin{enumerate}
    \item navigation: takes a location name and plans a path to the destination.
    \item object detection: takes an object name as an argument and detects the target object.
    \item pick: takes an object name as an argument and grasps the target object.
    \item place: takes a destination (in this study, the name of the mobile robot) as an argument and places the object.
    \item help: takes the content of the request as an argument and asks the operator for assistance.
\end{enumerate}
The prompt used for action planning was constructed based on our prior work~\cite{hasegawa2025spatial}.

The LLM outputs actions one at a time, and the action engine executes the program corresponding to each output action name.
After the action is executed, a binary value indicating success or failure is output, and the LLM determines the next action based on this result. 
The success or failure of each action is judged according to the following criteria. 
Navigation is judged by whether a path to the destination could be generated. 
Object detection is judged by whether a bounding box of the target object was detected in the image. 
Pick and place use two criteria. 
The first is whether inverse kinematics for the target object's position could be solved. 
The second is whether the object was actually grasped or placed, judged by image recognition based on a vision-language model.





\subsection{Remote Error Resolution and Feedback to Re-planning}

An action is judged to have failed according to the success/failure criteria described in Section~\ref{subsec:proposed_action_planning}. 
When this happens, the robot can either retry the action or issue the help skill to trigger RER. 
Both options are specified in the prompt, and the LLM decides between them.
Specifically, when navigation, object detection, picking, or placing fails, the mobile manipulator may retry the same or an alternative action, or request assistance from the operator via help. 
In particular, upon a failure in object detection or picking, the LLM instructs the operator to place the target object on top of the mobile robot.

When help is executed, the LLM generates a request describing the current failure and the problem the robot is facing, and asks the operator for assistance.
For example, when object detection fails, a request such as ``The target object could not be found, so please grasp it instead'' is generated.
At this point, action planning by the LLM is temporarily suspended, and the system transitions to a state in which the operator can teleoperate the robot. 
Using a teleoperation device, the operator controls the robot and performs the task on its behalf according to the request. 

After performing the task on the robot's behalf, the operator sends the details of what was done to the LLM as feedback. 
For example, the operator reports what was done in natural language, such as ``Remote-controlled Robot 2 grasped the Towel and placed it on top of Robot 1.''
Based on this report, the LLM updates its understanding of the situation and regenerates the subsequent action plan. 
In this way, our method integrates two elements: the resolution of failures through remote intervention, and the reflection of the intervention's outcome in the LLM's plan. 
This enables the task to continue even when a physical failure occurs that the robot cannot resolve without physical human intervention.




\section{Experiment}
\label{sec:experiment}
The purpose of this experiment is to verify whether feeding intervention outcomes back to the LLM enables the task to continue after a physical action failure is resolved through remote intervention. 
To this end, we compare task progress across three conditions: Teleop, the proposed system, and Auto. 
Task progress is measured as the number of pieces of trash collected within a fixed time limit.
In addition, using targets of differing collection difficulty, we examine the conditions under which intervention works effectively and those under which its effect is limited.
In this experiment, we use a mobile manipulator and a mobile robot in a space that emulates a domestic environment, where the task is to clear trash to a designated location.


\subsection{Task Definition}
\label{subsec:definition_of_tasks}

In this experiment, two robots perform a trash-collection task: they gather trash from the environment and transport it to a designated location within a time limit. 
The degree of human involvement differs across the three conditions described in Section~\ref{subsec:comaprison_methods}: in Auto the robots act fully autonomously, in the proposed system a human intervenes only when the robots request assistance, and in Teleop the participant operates the robots throughout.
The task is terminated under the following three conditions:
\begin{enumerate}
    \item All trash in the environment is collected and placed in the designated location within the time limit.
    \item The participant chooses to stop the task.
    \item The time limit (25 minutes) is exceeded.
\end{enumerate}

A total of six pieces of trash, belonging to four categories, were placed in the environment. 
The categories include paper waste, a partially filled bottle, an empty snack cup container, and an empty bottle. 
The developers classified these items into two levels of collection difficulty. 
Level~1 consists of objects that are easy to collect, including the partially filled bottle and the empty snack cup container. 
In contrast, Level~2 consists of objects that are difficult to collect, including the empty bottle and paper waste. 
For the empty bottle, pick-and-place operations may fail due to its tendency to tip over, while for paper waste, object detection may fail due to its irregular shape. 
By defining two levels of difficulty, the experiment assumes a scenario in which tasks that robots can handle autonomously and tasks that require human teleoperation are mixed. 
The placement of the trash is shown in Fig.~\ref{fig:exp_environment}; under the condition that one Level 1 and one Level 2 object are placed at each location, the specific positions are determined randomly.

The trash collection location is the Trash area in the environment shown in Fig.~\ref{fig:exp_environment}. 
In this experiment, a mobile robot (Kachaka) and a mobile manipulator (Fetch) are used. 
Kachaka is equipped with a cart on which a clear box is placed. 
Subjects collect trash by placing it into this clear box. 
Trash that is transported by Kachaka to the trash area within the time limit is considered ``collected.'' 
Only Kachaka is allowed to move to the Trash area.
In this experiment, the physical failures subject to RER are those of Fetch during object detection, picking, and placing. 
Kachaka is responsible only for transportation (navigation), and failures of Kachaka are not treated as targets for intervention.

Additional rules for the task are as follows. 
First, after the experiment begins, the developers do not provide subjects with any operational commands. 
Second, in cases such as robot collisions, as described below, the developers will restore the robots and return both robots to their initial positions (Fetch to the workspace and Kachaka to the trash area). 
However, the measurement of the time limit will not be paused.
\begin{enumerate}
    \item When the robots collide and the cart attached to Kachaka becomes detached.
    \item When Fetch's arm or base collides with a wall, desk, or Kachaka and exhibits clearly dangerous behavior.
\end{enumerate}
When the robots are returned to their initial positions, any objects that were in Kachaka's clear box are considered ``collected'' once Kachaka has returned to the Trash area.





\subsection{Experimental Setup}
\label{subsec:experiment_environment}

\begin{figure}[t]
        \centering
        \includegraphics[width=1.0\linewidth]{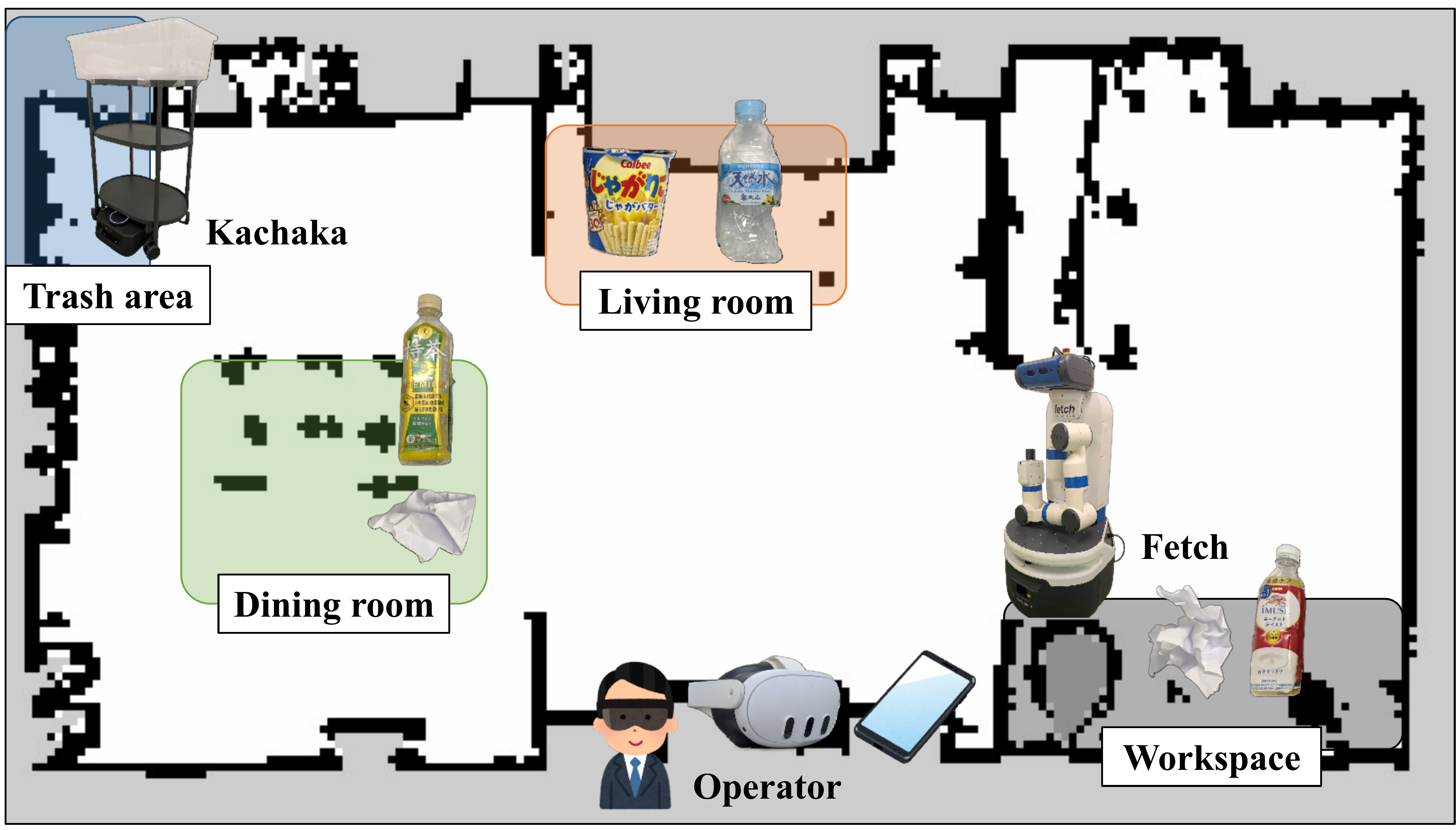}
        \caption{
            Experimental Environment and Object Placements.
        }
        \label{fig:exp_environment}
    \end{figure}

Fig.~\ref{fig:exp_environment} shows the environment used in this experiment.
The environment consists of four room regions predefined by the developer: the trash area, dining room, living room, and workspace.
At the start of the task, Kachaka was placed in the trash area, and Fetch in the workspace.

Each robot was given the environment map, the region of each room, and the navigation goal points as prior knowledge, together with information on the types and locations of the trash.
For Fetch's object detection, we used Detic~\cite{zhou2022detecting}.
We restricted the detection targets to the class names of the trash and used a pre-trained model~\footnote{https://github.com/facebookresearch/Detic/blob/main/configs/\\Detic\_LCOCOI21k\_CLIP\_R5021k\_640b32\_4x\_ft4x\_max-size.yaml}.

We used OpenAI's o3-2025-0416 for task decomposition and allocation, planning, and state recognition.
All other parameters were set to their default values.
Task decomposition and allocation were carried out in advance, before the experiment began.
For this step, we gave the LLM the instruction ``Collect all the trash in the rooms and transport it to the trash\_area.'' as the task-decomposition prompt.
The experimental PC ran a Docker-based development environment~\cite{el_hafi_software_2022} with Ubuntu 22.04 LTS and ROS Noetic.
The hardware configuration includes an Intel Core i9-12900H CPU and an NVIDIA GeForce RTX 3080 Ti GPU (12 GB).

In this experiment, participants operated both robots using teleoperation devices, according to the comparison method.
Fetch was operated with a Quest~3 running a teleoperation application developed in Unity, and Kachaka with a smartphone (Google Pixel 8a) running the Kachaka application provided by Preferred Robotics, Inc.

The experiment was conducted with a total of 12 subjects (8 males and 4 females)~\footnote{This study was conducted in accordance with the ``Ethical Guidelines for Research Involving Human Subjects at Ritsumeikan University.'' Based on the checklist of the Ritsumeikan University Research Ethics Review Committee, this study was determined not to meet the criteria requiring ethical review; therefore, formal ethical approval was not required. Written informed consent was obtained from all subjects.}. 
The subjects consisted of 11 undergraduate and graduate students and 1 researcher, with ages ranging from their 20s to 30s. 
Six of them had prior experience using an MR head-mounted display.






\subsection{Comparison Methods}
\label{subsec:comaprison_methods}

We prepared two methods for comparison.
First, Auto is a system in which the LLM performs task decomposition, assignment, and action planning, and automatically re-plans according to the success or failure of each action, without requesting human intervention even when a problem occurs.
Second, Teleop is a system in which the participant directly controls the two robots using a teleoperation device.

Next, we describe why we adopted Teleop and Auto as comparison methods.
Teleop was prepared as a human-operated reference condition indicating the task progress achievable when a human teleoperates the entire process.
Auto, on the other hand, was prepared to show that, when a physical failure such as object detection or grasping occurs, the autonomous baseline used in this study did not recover and task progress stalled.
In our system, the operator was instructed not to intervene unless explicitly requested by the robot, ensuring that assistance was provided strictly on demand.




\subsection{Evaluation Metrics}
\label{subsec:evaluation_metrics}
As the main evaluation metric, we used the number of pieces of trash the robots cleared by the end of the task (task progress). 
As a supplementary evaluation metric, we used the NASA-TLX~\cite{hart1988development}. 
The NASA-TLX is a method for subjectively assessing a participant's mental workload. 
We administered the NASA-TLX after each trial of each system.


\subsection{Experiment Flow}
\label{subsec:experiment_flow}
The experiment was conducted in person. 
First, subjects read and signed the informed consent form, and the developers explained the purpose, overview, and three robot operation systems (10 min). 
Then, all subjects completed the three conditions in the following order: Teleop (30 min practice + 25 min experiment + 5 min questionnaire), the proposed system (10 min practice + 25 min experiment + 5 min questionnaire), and Auto (10 min practice + 25 min experiment + 5 min questionnaire).


\subsection{Results}
\label{subsec:results}

\subsubsection{Task Progress}
\label{subsubsec:task_progress}
\begin{figure}[t]
        \centering
        \includegraphics[width=1.0\linewidth]{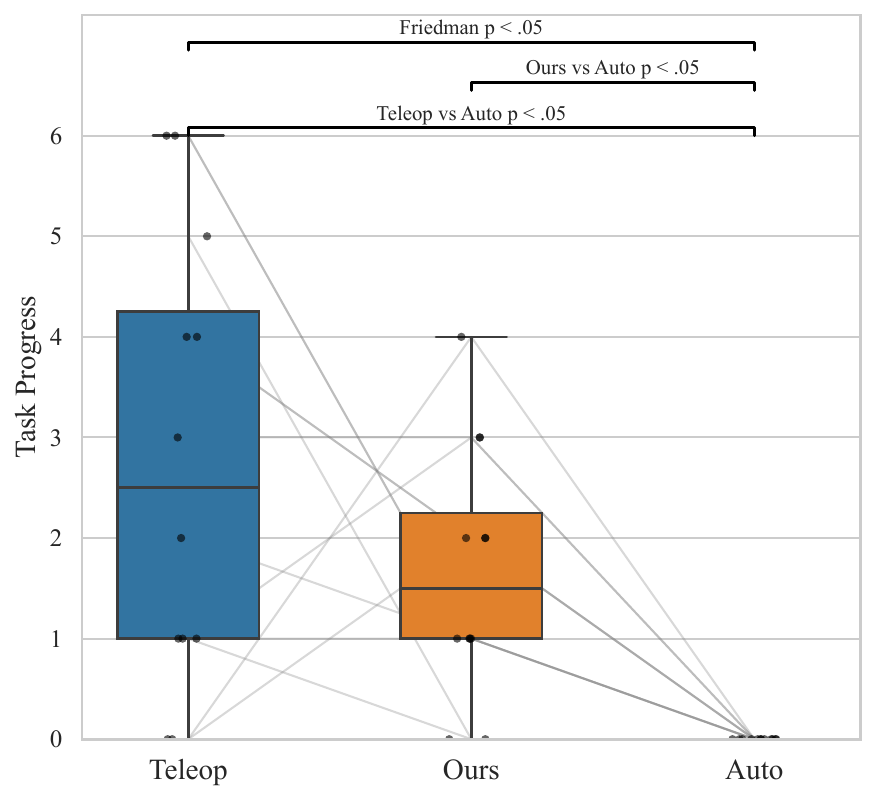}
        \caption{
            Distribution of task progress and results of statistical comparisons for each condition. The higher the task progress, the better (↑).
        }
        \label{fig:result_task_progress}
    \end{figure}

The results for task progress are shown in Fig.~\ref{fig:result_task_progress}.
A Friedman test (significance level: $0.05$) was conducted on task progress.
When a significant effect was observed, post hoc pairwise comparisons between two related conditions were performed using the Wilcoxon signed-rank test (two-tailed, significance level: $0.05$).
Furthermore, Holm's correction was applied to control the Type I error rate due to multiple comparisons.
The Friedman test revealed a significant effect ($p = 5.4 \times 10^{-4}$).
For the post hoc analysis, pairwise comparisons were conducted based on the following three hypotheses:
\begin{enumerate}
    \item H2-1: Teleop yields higher task progress than our method
    \item H2-2: Teleop yields higher task progress than Auto
    \item H2-3: Our method yields higher task progress than Auto
\end{enumerate}
The Wilcoxon signed-rank test with Holm correction revealed significant differences for H2-2 and H2-3 ($p = 0.0058$, $p = 0.0058$), whereas no significant difference was found for H2-1 ($p = 0.25$). 
The effect sizes ($r = |Z|/\sqrt{n}$) for the Wilcoxon signed-rank tests were $0.37$ (H2-1), $0.89$ (H2-2), and $0.90$ (H2-3), with Kendall's W $0.62$ for the Friedman test.

Next, the mean and standard deviation of task progress for each level of object collection difficulty are shown in Table~\ref{tab:result_task_progress_average}.
Table~\ref{tab:result_task_progress_average} breaks down task progress by object collection difficulty. 
For Level 1 objects, the gap between proposed system ($1.0 \pm 0.60$) and Teleop ($1.2 \pm 1.2$) was small, indicating that proposed system approached the Teleop reference. 
For Level 2 objects, however, the gap between proposed system ($0.66 \pm 0.77$) and Teleop ($1.5 \pm 1.0$) was larger. 
This contrast suggests that how closely proposed system approaches Teleop depends on the collection difficulty of the target object.

These results suggest statistically significant differences between Teleop and Auto, as well as between the proposed system and Auto.
Moreover, the large effect sizes ($0.89$ and $0.90$) suggest clear differences between the conditions.
A comparison of the mean values further shows that the proposed system improved performance compared to Auto, particularly for Level~1 objects.

\begin{table}[tb]
    \caption{
        Mean and standard deviation of task progress for each condition and each level of collection difficulty.
    }
    \label{tab:result_task_progress_average}
    \begin{center}
        \begin{tabularx}{1.0\linewidth}{lccc}
            \hline
            \textbf{Methods} & \textbf{Whole ↑} & \textbf{Level 1 Object ↑} & \textbf{Level 2 Object ↑} \\
            \hline
            Teleop & 2.7±2.2  & 1.2±1.2 & 1.5±1.0 \\
            Ours & 1.6±1.2  & 1.0±0.60 & 0.66±0.77 \\
            Auto & 0.0±0.0  & 0.0±0.0 & 0.0±0.0 \\
            \hline
        \end{tabularx}
    \end{center}
\end{table}

\subsubsection{NASA-TLX}
\label{subsubsec:nasa_tlx}

\begin{figure*}[t]
        \centering
        \includegraphics[width=0.9\linewidth]{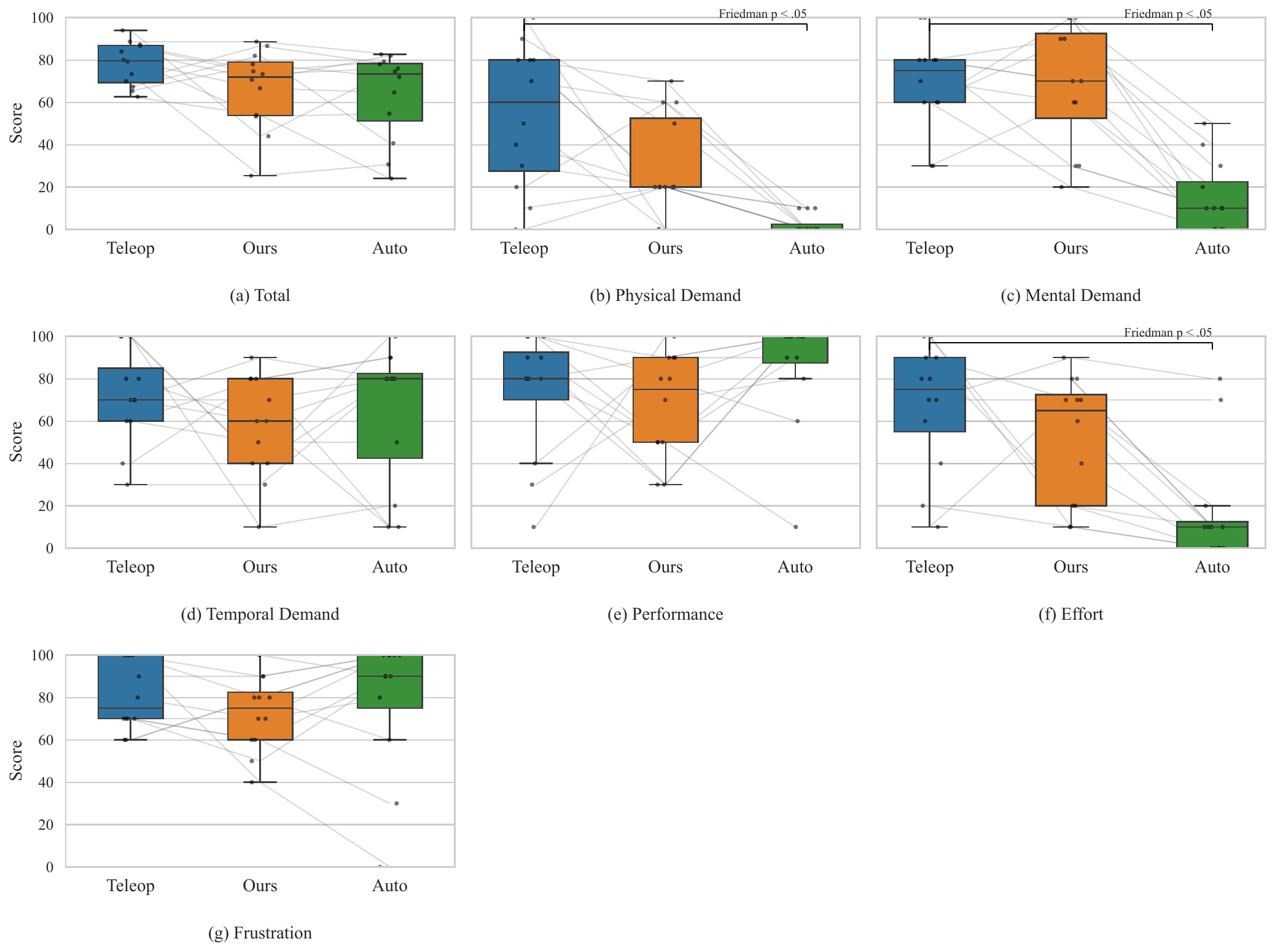}
        \caption{
            Fig.~\ref{fig:result_nasa_tlx}(a) shows the overall NASA-TLX score, while Fig.~\ref{fig:result_nasa_tlx}(b)--(g) show the scores for each subscale. The gray lines in each figure represent changes in scores across conditions for the same subject. In Fig.~\ref{fig:result_nasa_tlx}(b), (c), and (f), the Friedman test revealed significant differences among the three conditions ($p < 0.05$). A lower score indicates a better result (↓).
        }
        \label{fig:result_nasa_tlx}
    \end{figure*}

Figure~\ref{fig:result_nasa_tlx} shows the NASA-TLX results. 
The testing procedure is the same as for task progress. 
For the overall score (Fig.~\ref{fig:result_nasa_tlx}(a)), the Friedman test found no significant difference ($p = 0.064$), so no post-hoc comparison was conducted. 
We similarly performed a Friedman test on each subscale. 
We had formulated seven hypotheses in advance: that Teleop is higher than the proposed system on Physical, Mental, and Temporal Demand, Effort, and Frustration; that the proposed system is higher than Auto on Performance; and that Auto is higher on Frustration. 
Significant differences were found for Physical Demand, Mental Demand, and Effort ($p = 2.5 \times 10^{-4}$, $6.6 \times 10^{-5}$, and $6.0 \times 10^{-4}$, respectively).
However, for the predefined pairs, the Wilcoxon signed-rank test (with Holm correction) detected no significant difference in any case ($p = 0.21$, $0.89$, and $0.46$).


\section{Discussion}
\label{sec:discussion}

\begin{figure*}[t]
        \centering
        \includegraphics[width=1.0\linewidth]{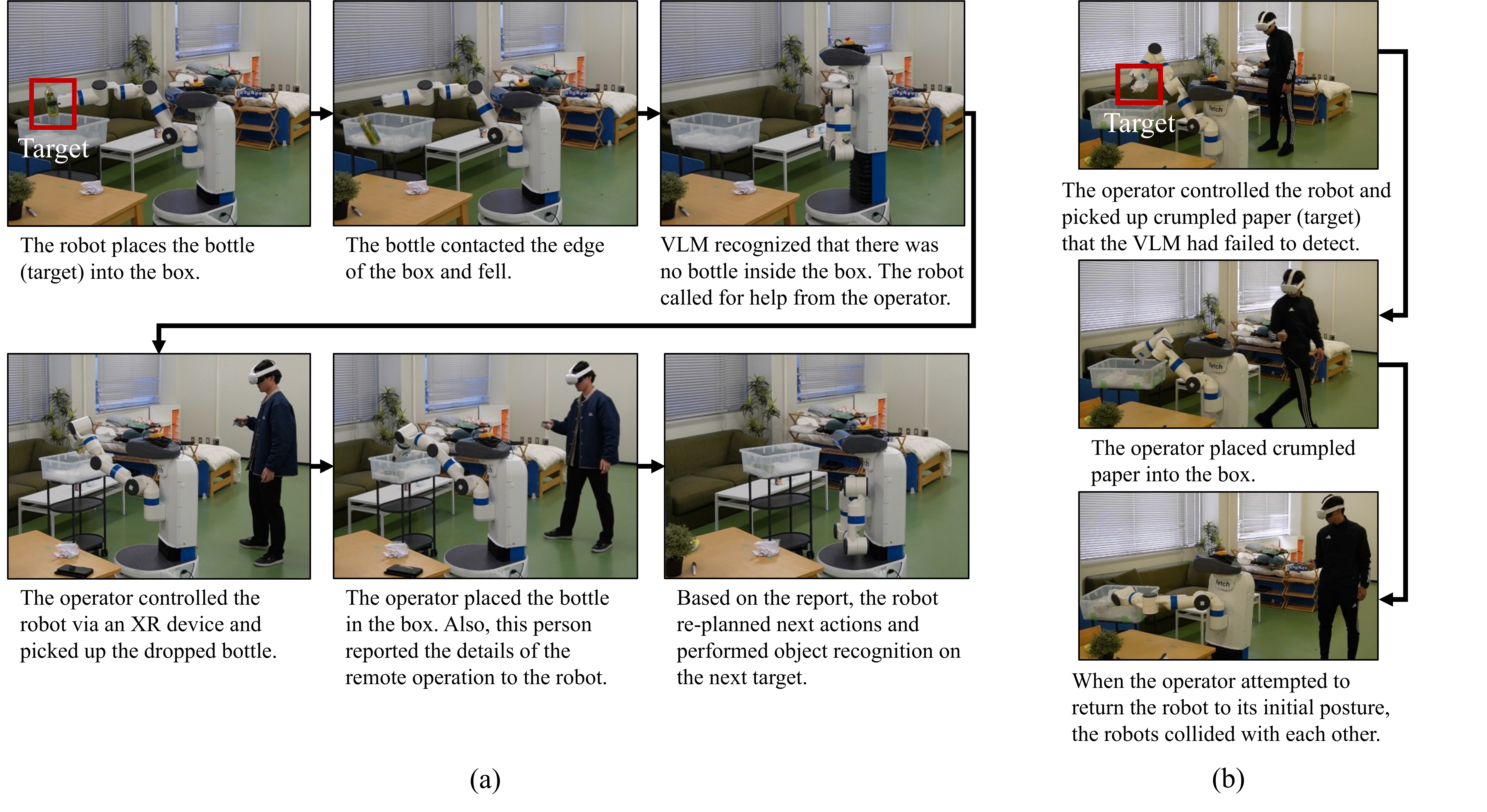}
        \caption{
            Demonstration of the proposed system in operation. (a) Operator performing remote error resolution for an object dropped by a robot. (b) Inter-robot collision due to operator remote control error.
        }
        \label{fig:result_flow}
    \end{figure*}

We first address task progress. 
In Auto, the robot could not recover from physical action failures and tended to repeat the same action. 
Two behaviors were typical. 
The robot repeatedly performed object detection without finding the target object. 
It also repeated object detection without noticing that it had dropped the object it was holding. 
Furthermore, in Auto, the robot often fell into a loop before reaching the transport stage. 
As a result, it never carried the object to the trash-collection point, and the trash was not counted as collected. 
In contrast, the proposed system could continue the task. 
When the robot recognized a failure, it requested assistance from the participant. We believe this contributed to the improvement in progress (Figure~\ref{fig:result_flow}~(a)).

Looking at the results by difficulty level, the proposed system's results for Level 1 were close to those of Teleop. 
We attribute this to the robot succeeding at object grasping even in autonomous mode. 
On the other hand, we identify two possible factors behind the drop in performance at Level 2.
The first concerns the participant's response to a request for assistance.
Rather than performing the task by teleoperation, the participant tried to resolve the situation with text-based instructions, resulting in insufficient communication of intent.
The second is that operation errors occurred when placing the object into the clear box. These errors led to collisions between the robots and a loss of time (Figure~\ref{fig:result_flow}~(b)).

For Teleop, qualitative comments suggested operational burden, including difficulty with arm control and eye fatigue from continuous VR use. 
For the proposed system, some participants commented that fatigue and human burden were reduced.





\section{Limitation}
\label{sec:limitation}
This study has several limitations. 
First, a different device is used for each teleoperated robot, so the operating environment may become complex. 
Second, Teleop is evaluated on the premise of an MR environment. 
As a result, we cannot separate whether the obtained results stem from the Teleop operation modality or from the MR interface.
Third, some factors could not be fully controlled: all participants performed the conditions in a fixed order (Teleop, ours, Auto), and practice times differed across conditions (30 min for Teleop, 10 min for the others) due to differences in operational proficiency required.
In addition, the failure rate experienced by each participant was not controlled.
We also did not separate whether failures stemmed from system-side factors or from the robot's capabilities.
Fourth, we do not deal in detail with the effects of erroneous failure judgments. 
Examples include issuing help when a success is misjudged as a failure, or failing to issue help because a failure was not detected.
Fifth, the comparison does not isolate the effect of feedback-based replanning from that of physical human intervention.

\section{Conclusion}
\label{sec:conclusion}
In this study, we built and evaluated a system for LLM-based robot action planning that continues a task by resolving physical action failures through remote human intervention.
When the robot fails at a physical action, the operator intervenes remotely to resolve the failure, and the outcome is fed back to the LLM and reflected in its plan.
In an evaluation using a real-world trash-collection task, we compared three conditions: fully autonomous (Auto), the proposed system, and teleoperation (Teleop). 
The results suggest that, in this case study, the proposed system improved task progress over Auto, and that for Level 1 objects the difference from Teleop was small.
On the other hand, for the NASA-TLX, which we used as a supplementary metric, no clear significant differences were found between conditions. 

In future work, we will analyze how the operator's response to a request for assistance (teleoperation or verbal instruction) affects task completion and workload. 
Moreover, we will work on integrating an immersive teleoperation environment that improves situational awareness during intervention.

\bibliographystyle{templates/IEEEtran}
\bibliography{root}

@article{zhai2025survey,
  title={{A Survey of Task Planning with Large Language Models}},
  author={Zhai, Wenshuo and others},
  journal={{Intelligent Computing}},
  volume={4},
  pages={0124},
  year={2025}
}

@article{ahn2022can,
  title={{Do As I Can, Not As I Say: Grounding Language in Robotic Affordances}},
  author={Ahn, Michael and others},
  journal={arXiv preprint arXiv:2204.01691},
  year={2022}
}

@article{tsushima2025task,
  title={{Task Planning for a Factory Robot using Large Language Model}},
  author={Tsushima, Yosuke and others},
  journal={{IEEE Robotics and Automation Letters}},
  volume={10},
  number={3},
  pages={2383--2390},
  year={2025}
}

@article{hasegawa2025spatial,
  title={{Spatial Concepts-Based Prompts With Large Language Models for Robot Action Planning}},
  author={Hasegawa, Shoichi and others},
  journal={{IEEE Access}},
  volume={13},
  pages={216937--216955},
  year={2025}
}

@inproceedings{liang2023code,
  title={{Code as Policies: Language Model Programs for Embodied Control}},
  author={Liang, Jacky and others},
  booktitle={{IEEE ICRA}},
  pages={9493--9500},
  year={2023}
}

@article{huang2022inner,
  title={{Inner Monologue: Embodied Reasoning through Planning with Language Models}},
  author={Huang, Wenlong and others},
  journal={arXiv preprint arXiv:2207.05608},
  year={2022}
}

@article{liu2023reflect,
  title={{REFLECT: Summarizing Robot Experiences for Failure Explanation and Correction}},
  author={Liu, Zeyi and others},
  journal={arXiv preprint arXiv:2306.15724},
  year={2023}
}

@article{ly2026inteliplan,
  title={{InteLiPlan: An Interactive Lightweight LLM-Based Planner for Domestic Robot Autonomy}},
  author={Ly, Kim Tien and others},
  journal={{IEEE Robotics and Automation Letters}},
  year={2026}
}

@article{liu2023llm,
  title={{LLM-based Human-robot Collaboration Framework for Manipulation Tasks}},
  author={Liu, Haokun and others},
  journal={arXiv preprint arXiv:2308.14972},
  year={2023}
}

@article{hagenow2025shared,
  title={{Shared Control/Autonomy: A Historical Perspective, Current Trends, and the Role of Generative AI}},
  author={Hagenow, Michael and others},
  journal={techrxiv preprint techrxiv:176617724.41163595},
  year={2025}
}

@inproceedings{kumar2025mixed,
  title={{Mixed Reality Outperforms Virtual Reality for Remote Error Resolution in Pick-and-Place Tasks}},
  author={Kumar, Advay and others},
  booktitle={{ACM/IEEE HRI}},
  pages={511--519},
  year={2025}
}

@inproceedings{wozniak2023happily,
  title={{Happily Error After: Framework Development and User Study for Correcting Robot Perception Errors in Virtual Reality}},
  author={Wozniak, Maciej K and others},
  booktitle={{IEEE RO-MAN}},
  pages={1573--1580},
  year={2023}
}

@inproceedings{kaipa2015resolving,
  title={{Resolving Automated Perception System Failures in Bin-Picking Tasks using Assistance from Remote Human Operators}},
  author={Kaipa, Krishnanand N and others},
  booktitle={{IEEE CASE}},
  pages={1453--1458},
  year={2015}
}

@article{murata2025multi,
  title={{Multi-Robot Task Planning for Multi-Object Retrieval Tasks with Distributed On-Site Knowledge via Large Language Models}},
  author={Murata, Kento and others},
  journal={{Artificial Life and Robotics}},
  year={2026}
}

@inproceedings{shirasaka2024self,
  title={{Self-Recovery Prompting: Promptable General Purpose Service Robot System with Foundation Models and Self-Recovery}},
  author={Shirasaka, Mimo and others},
  booktitle={{IEEE ICRA}},
  pages={17395--17402},
  year={2024}
}

@inproceedings{Heger2006sliding,
  title={{Sliding Autonomy for Complex Coordinated Multi-Robot Tasks: Analysis \& Experiments}},
  author={Heger, Frederik and others},
  booktitle={{RSS}},
  year={2006}
}

@article{liu2025robot,
  title={{Robot Learning on the Job: Human-in-the-loop Autonomy and Learning during Deployment}},
  author={Liu, Huihan and others},
  journal={{The International Journal of Robotics Research}},
  volume={44},
  number={10-11},
  pages={1727--1742},
  year={2025}
}

@article{el_hafi_software_2022,
	title = {Software {Development} {Environment} for {Collaborative} {Research} {Workflow} in {Robotic} {System} {Integration}},
	volume = {36},
	number = {11},
	journal = {{Advanced Robotics}},
	author = {El Hafi, Lotfi and others},
	year = {2022},
	pages = {533--547}
}

@inproceedings{
    zhou2022detecting,
    title={{Detecting Twenty-Thousand Classes Using Image-Level Supervision}},
    author={Zhou, Xingyi and others},
    booktitle={{ECCV}},
    pages={350--368},
    year={2022}
}

@incollection{hart1988development,
  title={{Development of NASA-TLX (Task Load Index): Results of Empirical and Theoretical Research}},
  author={Hart, Sandra G and others},
  booktitle={{Advances in Psychology}},
  volume={52},
  pages={139--183},
  year={1988}
}

\end{document}